\pdfoutput=1

\documentclass[11pt]{article}

\usepackage{EMNLP2022}

\usepackage{times}
\usepackage{latexsym}

\usepackage{amsmath}
\usepackage{algpseudocode}
\usepackage{algorithm}
\usepackage{mathtools}

\DeclareCaptionLabelFormat{noname}{#2}
\usepackage{tikz}
\usepackage{amssymb}
\usepackage{environ}
\usepackage{bbm}

\usepackage{amsmath}

\DeclareMathOperator{\len}{len}
\DeclareMathOperator{\append}{append}

\makeatletter
\newsavebox{\measure@tikzpicture}
\NewEnviron{scaletikzpicturetowidth}[1]{%
  \def\tikz@width{#1}%
  \def\tikzscale{1}\begin{lrbox}{\measure@tikzpicture}%
  \BODY
  \end{lrbox}%
  \pgfmathparse{#1/\wd\measure@tikzpicture}%
  \edef\tikzscale{\pgfmathresult}%
  \BODY
}
\makeatother

\usepackage[T1]{fontenc}

\usepackage[utf8]{inputenc}

\usepackage{microtype}

\usepackage{inconsolata}
\usepackage{tipa}
\usepackage{subfig}

\usepackage{graphicx}

\DeclareMathOperator*{\argmax}{arg\,max}

\title{Neural Unsupervised Reconstruction of Protolanguage Word Forms}

\author{Andre He \qquad Nicholas Tomlin \qquad Dan Klein \\
  Computer Science Division, University of California, Berkeley \\
  \texttt{\{andre.he, nicholas\_tomlin, klein\}@berkeley.edu}}

\begin{document}
\maketitle
\begin{abstract}
We present a state-of-the-art neural approach to the unsupervised reconstruction of ancient word forms.
Previous work in this domain used expectation-maximization to predict simple phonological changes between ancient word forms and their cognates in modern languages. 
We extend this work with neural models that can capture more complicated phonological and morphological changes.
At the same time, we preserve the inductive biases from classical methods by building monotonic alignment constraints into the model and deliberately underfitting during the maximization step.
We evaluate our performance on the task of reconstructing Latin from a dataset of cognates across five Romance languages, achieving a notable reduction in edit distance from the target word forms compared to previous methods.

\end{abstract}
\section{Introduction}

Research has shown that groups of languages can often be traced back to a common ancestor, or a \textit{protolanguage}, which has evolved and branched out over time to produce its modern descendants. Words in protolanguages undergo sound changes to produce their corresponding forms in modern languages. We call words in different languages with a common proto-word ancestor \textit{cognates}. The study of cognate sets can reveal patterns of phonological change, but their proto-words are often undocumented \citep{campbell2013historical, Hock+2021}.

\begin{figure*}
\centering 
\includegraphics[width=1.0\textwidth]{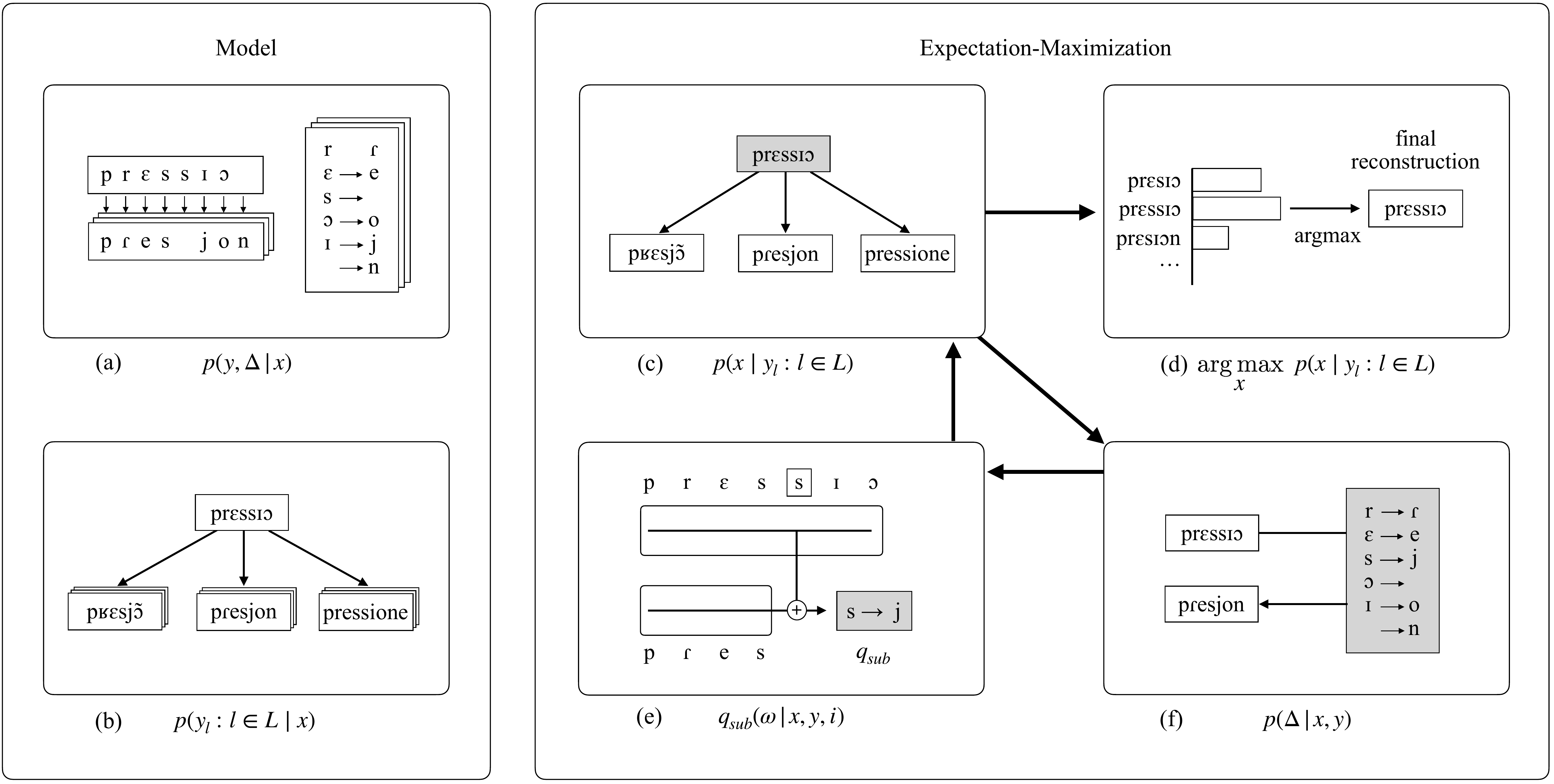}
\caption{Overview of our paper. (a) We model the evolution of word forms as a generative process which applies many character-level edits to the ancestral form, producing a distribution over the output word form $y$ and edit sequence $\Delta$. (b) Using a dynamic program, we can compute the distribution over output words, $p(y \mid x)$. We model this for every language branch $l \in L$. (c) Our method uses EM to infer ancestral forms. For the E-step, we want to sample from the posterior distribution, where $y$ is observed but $x$ is not. (f) With samples from the previous step fixed, we use another dynamic program to compute expected edit counts. (e) In the M-step, we use these edit counts to train our character-level edit models $q$, parameterized as recurrent neural networks. $q$ determines the edit probabilities in (c) and thus influences the next round of samples. (d) After several EM iterations, we take the maximum likelihood word forms as the final reconstructions.}
\label{fig:overview}
\end{figure*}

To reconstruct ancient word forms, linguists use the comparative method, which compares individual features of words in modern languages to their corresponding forms in hypothesized reconstructions of the protolanguage. Past work has demonstrated the possibility of automating this manual procedure \citep{durham1969application, eastlack1977iberochange, lowe1994reconstruction, covington-1998-alignment-multiple, kondrak2002algorithms}.
For example, \citet{bouchard-etal-2007-probabilistic, NIPS2007_7ce3284b} developed probabilistic models of phonological change and used them to learn reconstructions of Latin based on a dataset of Romance languages, and  \citet{bouchard-cote-etal-2009-improved, doi:10.1073/pnas.1204678110} extended their method to a large scale dataset of Austronesian languages \citep{greenhill2008austronesian}.

Nevertheless, previous approaches to computational protolanguage reconstruction have mainly considered simple rules of phonological change. In previous works, phonological change is modeled applying a sequence of phoneme-level edits to the ancestral form. Although this can capture many regular sound changes such as lenitions, epentheses, and elisions \citep{doi:10.1073/pnas.1204678110}, these edits are typically conditioned only on adjacent phonemes and lack more general context-sensitivity. Phonological effects such as dissimilation \citep{bye2011dissimilation}, vowel harmony \citep{nevins2010locality}, syllabic stress \citep{sen2012reconstructing}, pre-cluster shortening \citep{yip1987english}, trysyllabic laxing \citep{mohanan1982lexical}, and homorganic lengthening \citep{wena1998functional}, as well as many non-phonological aspects of language change \citep{fisiak2011historical}, are all frequently dependent on non-local contexts.
However, it is difficult to extend existing multinomial \citep{bouchard-etal-2007-probabilistic} and log-linear \citep{NIPS2007_7ce3284b, bouchard-cote-etal-2009-improved, doi:10.1073/pnas.1204678110} models to handle more complex conditioning environments.

Motivated by these challenges, our work is the first to use neural models for unsupervised reconstruction. 
Ancestral word forms and model parameters in previous unsupervised approaches are typically learned using expectation-maximization \citep[e.g.,][]{bouchard-etal-2007-probabilistic}. 
In applying neural methods to protolanguage reconstruction, we identify a problem in which the EM objective becomes degenerate under highly expressive models.
In particular, we find that neural models are able to express not just complex phonological changes, but also \textit{inconsistent} ones (i.e., predicting vastly different edits in similar contexts), undermining their ability to distinguish between good and bad hypotheses. 
From a linguistic perspective, phonological change should exhibit regularities due to the constraints of the human articulatory and cognitive faculties \citep{kiparsky1965phonological}, so we build a bias towards regular changes into our method by using a specialized model architecture and learning algorithm. We outline our approach in Figure~\ref{fig:overview}.

Our work enables neural models to effectively learn reconstructions under expectation-maximization.
In Section~\ref{sec:model}, we describe a specialized neural architecture with monotonic alignment constraints. In Section~\ref{sec:underfitting}, we motivate training deliberately underfitted models. Then, in Section~\ref{sec:experiments}, we conduct experiments and show a significant improvement over the previously best performing method. Finally, we conduct ablation experiments and attribute the improvement to (1) the ability to model longer contexts and (2) a training process that is well-regularized for learning under EM.

\section{Related Work}
\label{sec:relatedwork}

Our work directly extends a series of previous approaches to unsupervised protolanguage reconstruction that model the probabilities of phoneme-level edits from ancestral forms to their descendants \citep{bouchard-etal-2007-probabilistic, NIPS2007_7ce3284b, bouchard-cote-etal-2009-improved, doi:10.1073/pnas.1204678110}. 
These edits include substitutions, insertions, and deletions, with probabilities conditioned on the local context. The edit model parameters and unknown ancestral forms are jointly learned with expectation-maximization. 
The main difference between models in previous work is in parameterization and conditioning: \citet{bouchard-etal-2007-probabilistic} used a multinomial model conditioned on immediate neighbors of the edited phoneme; \citet{NIPS2007_7ce3284b} used a featurized log-linear model with similar conditioning; and \citet{bouchard-cote-etal-2009-improved} introduced markedness features that condition on the previous output phoneme. \citet{bouchard-cote-etal-2009-improved} also shared parameters across branches so that the models could learn global patterns. \citet{doi:10.1073/pnas.1204678110} used essentially the same model but ran more comprehensive experiments on a larger dataset. 

Since the expectation step of EM is intractable over a space of strings, past work resort to a Monte-Carlo EM algorithm where the likelihood is optimized with respect to sample ancestral forms. However, this sampling step is still the bottleneck of the method as it requires computing data likelihoods for a large set of proposed reconstructions. \citet{bouchard-etal-2007-probabilistic} proposed a single-sequence resampling method, but this approach propagated information too slowly in deep phylogenetic trees, so \citet{bouchard-cote-etal-2009-improved} replaced it with a method known as ancestry resampling \citep{NIPS2008_1651cf0d}. This method samples an entire ancestry at a time, defined as a thin slice of aligned substrings across the tree that are believed to have descended from a common substring of the proto-word. 
Changes since the \citet{bouchard-cote-etal-2009-improved} work, including shared parameters and ancestry resampling, are primarily concerned with reconstruction in large phylogenetic trees. While they improve reconstruction quality drastically on the Austronesian dataset, these modifications did not bring a statistically significant improvement on the task of reconstructing Latin from a family of Romance languages \citep{bouchard-cote-etal-2009-improved}. This is likely due to the Romance family consisting of a shallow tree of a few languages, where the main concern is learning more complex changes on each branch. Therefore, in this work we compare our model to that of \citet{bouchard-cote-etal-2009-improved} but keep the single sequence resampling method from \citet{bouchard-etal-2007-probabilistic}. 

Previous work also exists on the related task of supervised protolanguage reconstruction. This is an easier task because models can be directly trained on gold reconstructions. \citet{meloni-etal-2021-ab} trained a GRU-based encoder-decoder architecture on cognates from a family of five Romance languages to predict their Latin ancestors and achieved low error from the ground truth. Another similar supervised character-level sequence-to-sequence task is the prediction of morphological inflection. Recent work on this task by \citet{DBLP:journals/corr/AharoniG16} improved output quality from out-of-the-box encoder-decoders by modifying the architecture to use hard monotonic attention, constraining the decoder’s attention to obey left-to-right alignments between source and target strings. In our work, we find that character-level alignments is also an important inductive bias for unsupervised reconstruction. 

\section{Task Description}
In the task of protolanguage reconstruction, our goal is to predict the IPA representation of a list of words in an ancestral language. We have access to their cognates in several modern languages, which we believe to have evolved from their ancestral forms via regular sound changes. Following prior work \citep[e.g.,][]{bouchard-etal-2007-probabilistic, NIPS2007_7ce3284b}, we do not observe any ancestral forms directly but assume access to a simple (phoneme-level) bigram language model of the protolanguage.
We evaluate the method by computing the average edit distance between the model's outputs and gold reconstructions by human experts. 

Concretely, let $\Sigma$ be the set of IPA phonemes. We consider word forms that are strings of phonemes in the set $\Sigma^*$. We assume there to be a collection of cognate sets $C$ across a set of modern languages $L$. A cognate set $c \in C$ is in the form $\{y_l^c: l \in L\}$, consisting of one word form for each language $l$. We assume that cognates descend from a common proto-word $x^c$ through language-specific edit probabilities $p_l(y_l \mid x)$.
Initially, neither the ancestral forms $\{x^c: c \in C\}$ nor the edit probabilities $\{p_l(y_l \mid x), l \in L\}$ are known, and we wish to infer them from just the observed cognate sets $C$ and a bigram model prior $p(x)$.

\section{Dataset}
\label{sec:dataset}
In our setup, $L$ consists of four Romance languages, and Latin is the protolanguage. We use the dataset from \citet{meloni-etal-2021-ab}, which is a revision of the dataset of \citet{dinu-ciobanu-2014-building} with the addition of cognates scraped from Wiktionary. The original dataset contains 8799 cognates in Latin, Italian, Spanish, Portuguese, French, and Romanian. We follow \citet{meloni-etal-2021-ab} and use the \texttt{espeak} library\footnote{\url{https://github.com/espeak-ng/espeak-ng}} to convert the word forms from orthography into their IPA transcriptions. To keep the dataset consistent with the closest prior work on the unsupervised reconstruction of Latin \citep{bouchard-cote-etal-2009-improved}, we remove vowel length indicators and suprasegmental features, keep only full cognate sets, and drop the Romanian word forms. The resulting dataset has an order of magnitude more data ($|C|=3214$ vs. $586$) but is otherwise very similar. We show example cognate sets in the appendix.

\section{Model}
\label{sec:model}
In this section, we describe our overall model of the evolution of word forms. We organize the languages into a flat tree, with Latin at the root and the other Romance languages $l \in L$ as leaves. Following \citet{bouchard-etal-2007-probabilistic}, our overall model is generative and describes the production of all word forms in the tree. 
Proto-words are first generated at the root according to a prior $p(x)$, which is specified as a bigram language model of Latin. These forms are then rewritten into their modern counterparts at the leaves through branch-specific edit models denoted $p_l(y_l \mid x)$.

In using neural networks to parameterize the edit models, our preliminary experiments suggested that standard encoder-decoder architectures are unlikely to learn reasonable hypotheses when trained with expectation maximization. We identified this as a degeneracy problem: the space of possible changes expressible by these models is too large for unsupervised reconstruction to be feasible. Hence, we enforce the inductive bias that the output word form is produced from a sequence of local edits; these edits are conditioned on the global context so that the overall model is still highly flexible.  

In particular, to construct the word-level edit models, we first use a neural network to model context-sensitive, character-level edits. We then construct the word-level distribution via an iterative procedure that samples many character-level edits. We describe these components in the reverse order as the character-level distributions are clearer in the context of the edit process: in Section~\ref{sec:editproc}, we describe the edit process, while Section~\ref{sec:editmodel} details how we model the underlying character-level edits.

\subsection{Word-Level Edit Process}
\label{sec:editproc}
Given an ancestral form, our model transduces the input string from left to right and chooses edits to apply to each character. For a given character, the model first predicts a substitution outcome to replace it with. A special outcome is to delete the character, in which case the model skips to editing the next character. Otherwise, the model enters an insertion phase, where it sequentially inserts characters until predicting a special token that ends the insertion phase. After a deletion or end-of-insertion token occurs, the model moves on to editing the next input character. We describe the generative process in pseudocode in Figure~\ref{fig:pseudocode}.

\algdef{SE}[DOWHILE]{Do}{doWhile}{\algorithmicdo}[1]{\algorithmicwhile\ #1}%
\algnewcommand{\LineComment}[1]{\State \(\triangleright\) #1}
\begin{figure}[h]
\textbf{Input}: An ancestral word form $x$ \\
\textbf{Output}: A modern form $y$ and lists of edits $\Delta$ 
\begin{algorithmic}[1]
\Function{Edit}{$x$}
    \State $y' \gets [ \ ]$ 
    \State $\Delta \gets [ \ ]$ 
    \For {$j = 1, \dots, \len(x)$}            
        \LineComment{Sample substitution outcome}
        \State Sample $\omega \sim q_{\text{sub}}(\cdot \mid x, i, y')$  
        \State $\Delta.\append((\text{sub}, \omega, x, i, y'))$
        \If {$\omega \neq \text{<del>}$} 
            \Do 
                \State $y'.\append(\omega)$
                \LineComment{Sample insertion outcomes}    
                \State Sample $\omega \sim q_{\text{ins}}(\cdot \mid x, i, y')$     
                \State $\Delta.\append((\text{ins}, \omega, x, i, y'))$
            \doWhile {$\omega \neq \text{<end>}$}
        \EndIf
    \EndFor
    \State \Return $y' \text{ as } y$, $\Delta$    
\EndFunction
\end{algorithmic}
\caption{Pseudocode describing the generative process behind $p(y, \Delta \mid x)$. Each input character is potentially deleted or substituted, with zero or more characters inserted afterwards. The probabilities of edits are specified by the character-level edit models $q_{\text{sub}}$ and $q_{\text{ins}}$ (\ref{sec:editmodel}). Each edit in the list $\Delta$ is represented as a tuple $(\text{op}, \omega, x, i, y')$, where $\text{op} \in \{\text{sub}, \text{ins}\}$, $\omega \in \Sigma$, and $(x, i, y')$ make up the context of the edit.} 
\label{fig:pseudocode}
\end{figure}

The models $q_{\text{sub}}$ and $q_{\text{ins}}$ are our character-level edit models, and they control the outcome of substitutions and insertions, conditioned on $x$, the input string, $i$, the index of the current character, and $y'$, the output prefix generated so far. The distribution $q_{\text{sub}}$ is defined over $\Sigma \cup \{\text{<del>}\}$ and $q_{\text{ins}}$ is defined over $\Sigma \cup \{\text{<end>}\}$. Models in previous work can be seen as special cases of this framework, but they are limited to a 1-character input window around the current index, $x[i-1:i+1]$, and a 1-character history in the output, $y'[-1]$ \citep[e.g., in][]{bouchard-cote-etal-2009-improved}.

The generative process defines a distribution $p(y, \Delta \mid x)$ over the resulting modern form and edit sequences. But what we actually want to model is the distribution over modern word forms themselves -- for this purpose, we use a dynamic program to sum over valid edit sequences: $$p(y\mid x) = \sum_{\Delta} p(y, \Delta \mid x)$$
where $\Delta$ represents edits from $x$ into $y$ (see Appendix~\ref{appendix:fdp} for more details). The edit procedure, character-level models, and dynamic program together give a conditional distribution over modern forms. Note that we have one such model for each language branch. 

\subsection{Character-Level Model}
\label{sec:editmodel}

We now describe the architecture behind $q_{\text{sub}}$ and $q_{\text{ins}}$, which model the distribution over character-level edits conditioned on the appropriate inputs. Our model leverages the entire input context and output history by using recurrent neural networks. The input string $x$ is encoded with a bidirectional LSTM, and we take the embedding at the current index, denoted $h(x)[i]$. The output prefix $y'$ is encoded with a unidirectional LSTM, and we take the final embedding, which we call $g(y')[-1]$. The sum of these two embeddings $h(x)[i] + g(y')[-1]$ encodes the full context of an edit -- we apply two different classification heads to predict the substitution distribution $q_{\text{sub}}$ and the insertion distribution $q_{\text{ins}}$.
We note that the flow of information in our model is similar to the hard monotonic attention model of \citet{DBLP:journals/corr/AharoniG16}, which used an encoder-decoder architecture with a hard left-to-right attention constraint for supervised learning of morphological inflections. Figure~\ref{fig:arch} illustrates the model architecture with an example prediction.

\tikzstyle{cell} = [rectangle, draw, rounded corners, minimum height=30.0, minimum width=12.5]

\begin{figure*}
\centering 
\begin{scaletikzpicturetowidth}{0.9\textwidth}
\begin{tikzpicture}[scale=\tikzscale]

\node [cell] (e1) at (0.5, 2){};
\node [cell] (e2) at (2.5, 2){};
\node [cell] (e3) at (4.5, 2){};
\node [cell] (e4) at (6.5, 2){};
\node [cell] (e5) at (8.5, 2){};
\node [cell] (e6) at (10.5, 2){};

\node [anchor=base] at (-1.5, 4) {$x:$};
\node [anchor=base] at (0.5, 4) {p};
\node [anchor=base] at (0.5, 1.8) {$1$};
\node [anchor=base] at (2.5, 4) {r};
\node [anchor=base] at (2.5, 1.8) {$2$};
\node [anchor=base] at (4.5, 4) {\textipa{E}};
\node [anchor=base] at (4.5, 1.8) {$...$};
\node [anchor=base] at (6.5, 4) {s};
\node [anchor=base] at (6.5, 1.8) {$...$};
\node [anchor=base] at (8.5, 4) {\textipa{I}};
\node [anchor=base] at (8.5, 1.8) {$i$};
\node [anchor=base] at (10.5, 4) {\textipa{O}};
\node [anchor=base] at (10.5, 1.8) {$...$};
\draw [anchor=base] (8, 3.6) rectangle (9, 4.6){};

\node [cell] (d1) at (0.5, -2){};
\node [cell] (d2) at (2.5, -2){};
\node [cell] (d3) at (4.5, -2){};
\node [cell] (d4) at (6.5, -2){};

\node [anchor=base] at (-1.5, -4) {$y':$};
\node [anchor=base] at (0.5, -4) {p};
\node [anchor=base] at (2.5, -4) {e};
\node [anchor=base] at (4.5, -4) {s};
\node [anchor=base] at (6.5, -4) {s};
\draw [anchor=base, dashed] (8, -3.4) rectangle (9, -4.4){};

\draw [<->] (e1.east) -- (e2.west);
\draw [<->] (e2.east) -- (e3.west);
\draw [<->] (e3.east) -- (e4.west);
\draw [<->] (e4.east) -- (e5.west);
\draw [<->] (e5.east) -- (e6.west);
\draw [->] (d1.east) -- (d2.west);
\draw [->] (d2.east) -- (d3.west);
\draw [->] (d3.east) -- (d4.west);

\node [circle, draw](add) at (8.5, -0.5) {+}; 
\draw (d4.north) -- (6.5, -0.5);
\draw [->](6.5, -0.5) -- (add.west);
\draw [->](e5.south) -- (add.north);
\draw [->](add.east) -- (12.5, -0.5);

\draw [] (12.5, -3.0) rectangle (17.5, 3.0);
\node at (15, 2){$q_{\text{sub}}$};
\node at (14, 1){i};
\node at (16, 1){0.8};
\node at (14, 0){\textipa{I}};
\node at (16, 0){0.1};
\node at (14, -1){\textless del\textgreater};
\node at (16, -1){0.05};
\node at (15, -2){\ldots};

\draw [] (17.5, -3.0) rectangle (22.5, 3.0);
\node at (20, 2){$q_{\text{ins}}$};
\node at (19, 1){\textless end\textgreater};
\node at (21, 1){0.5};
\node at (19, 0){e};
\node at (21, 0){0.4};
\node at (19, -1){j};
\node at (21, -1){0.05};
\node at (20, -2){\ldots};

\node [align=center] at (-3, 2){Bidirectional \\ LSTM $h$};
\node [align=center] at (-3, -2){Unidirectional \\ LSTM $g$};
\end{tikzpicture}
\end{scaletikzpicturetowidth}
\caption{Architecture diagram of the character-level edit model, denoted $q_{\text{sub/ins}}(\omega \mid x, i, y')$. The distribution of outcomes is dependent on both the input string and output history. Here our model is shown predicting edits for \textit{\textipa{I}} when the input is \textit{\textipa{prEsIO}} and the current output is \textit{pess}.  The model predicts substitutions if the input character \textit{\textipa{I}} has not produced any outputs yet; otherwise it predicts characters to insert. Note that deletion \text{<del>} and end-of-insertion \text{<end>} are special outcomes of substitution and insertion. }
\label{fig:arch}
\end{figure*}
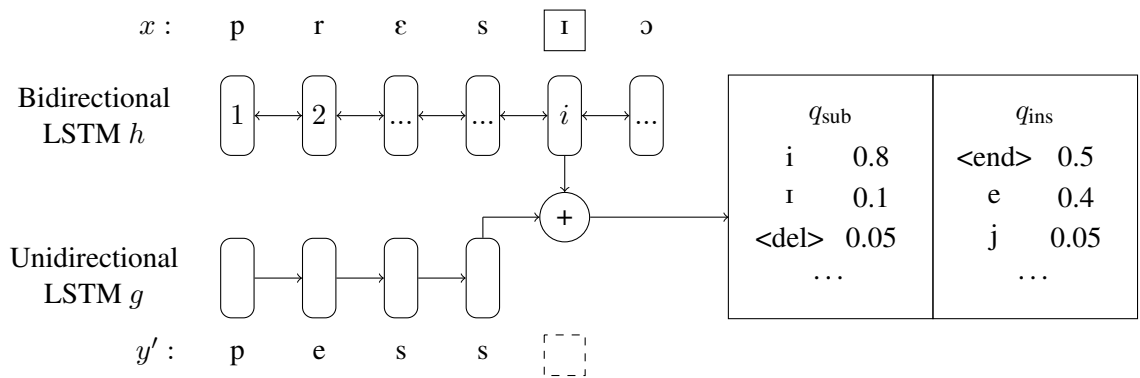

\section{Learning Algorithm}
The problem of unsupervised reconstruction is to infer the ancestral word forms $\{x^c: c\in C\}$ and edit models $\{p_l(y_l\mid x): l \in L\}$ when given the modern cognates $\{y_l^c: c\in C, l\in L\}$. 
We use a Monte-Carlo EM algorithm to learn the reconstructions and model parameters.
During the E-step, we seek to sample ancestral forms from the current model's posterior distribution, conditioned on observed modern forms; during the M-step, we train the edit models to maximize the likelihood of these samples. We alternate between the E and M steps for several iterations; then in the final round, instead of sampling, we take the maximum likelihood strings as predicted reconstructions. 

\subsection{Sampling Step}
The goal of the E-step is to sample ancestral forms from the current model’s posterior distribution, $p(x^c \mid \{y^c_l, l \in L\} )$. In general, this distribution cannot be computed directly; but for given samples of $x$, we can compute a value that is proportional to their posterior probability.  
At the beginning of an E-step, we have the current edit models $\{p_l(y_l\mid x): l\in L\}$, observed modern forms $\{y_l: l\in L\}$, and the ancestral form prior $p(x)$. For a given ancestral word form $x$, we can use Bayes' rule to compute a joint probability that is proportional to its posterior probability (our model assumes conditionally independent branches):

\begin{equation}
\begin{split}
p(x & \mid \{y_l, l \in L\}) \\ 
 & = \frac{p(x, \{y_l, l \in L\})}{p(\{y_l, l \in L\})} \\
 & \propto p(x, \{y_l, l \in L\}) \\
 & = p(x) \prod_{l \in L} p(y_l \mid x) 
\end{split}
\end{equation}
Following previous work, we use Metropolis-Hastings to sample from the posterior distribution without computing the normalization factor. We iteratively replace the current word form $x$ with a candidate drawn from a set of proposals, with probability proportional to the joint probability computed above. We repeat this process for each cognate set to obtain a set of sample ancestral forms $\{x^c: c \in C\}$.

During Metropolis-Hastings, the cylindrical proposal strategy in \citet{NIPS2008_1651cf0d} considers candidates within a 1-edit-distance ball of the current sample, but this strategy is inefficient since the number of proposals is scales linearly with both the string length and vocabulary size, and the sample changes by only one edit per round. We develop a new proposal strategy which exploits the low edit distance between cognates. Our approach considers all strings on a minimum edit path from the current sample to a modern form. This allows the current sample to move many steps at a time towards one of its modern cognates. 
See Figure~\ref{fig:mineditpath} in the appendix for an illustration.

\subsection{Maximization Step}
\label{sec:mstep}
With samples from the previous step $\{x^c: c \in C\}$ fixed, the goal of the M-step is to train our edit models to maximize data likelihood. The models on each branch are independent, so we train them separately. For each branch $l$, we wish to optimize
$$\sum_{c \in C} p(y_l^c \mid x^c)$$   
This is a standard sequence-to-sequence training objective, where the training set is simply ancestral forms $x^c$ from the E-step and modern forms $y_l^c$ from the dataset. However, since we do not directly model the conditional distribution of output strings (\ref{sec:editmodel}), we need the underlying edit sequences to train our character-level edit models $q_{\text{sub}}$ and $q_{\text{ins}}$. 

Given an input-output pair $x$ and $y$, we compute the probabilities of underlying edits using a dynamic program similar to the forward-backward algorithm for HMMs (see \ref{appendix:bdp} for more details). Concretely, for each possible substitution $(\text{sub}, \omega, x, i, y')$ defined as in Figure~\ref{fig:pseudocode}, the dynamic program computes
$$p((\text{sub}, \omega, x, i, y') \in \Delta \mid x, y)$$
which is the probability of the edit occurring, conditioned on the initial and resultant strings. We average over cognate pairs to obtain $p((\text{sub}, \omega, x, i, y') \in \Delta)$ and train the substitution model $q_{\text{sub}}(\omega \mid x, i, y')$ to fit this distribution. We compute insertion probabilities and train the insertion model in the same way. 

We bootstrap the neural models $q_{\text{sub}}$ and $q_{\text{ins}}$ by using samples from the classical method. Before the first maximization step, we train a model from \citet{bouchard-cote-etal-2009-improved} for three EM iterations. We use samples from the model to compute the first round of edit probabilities. Once the neural model is trained on these probabilities, we no longer rely on the classical model.
Note that this does not bias the comparison in Section~\ref{sec:comparison} in our favor because the classical models reach peak performance in less than five EM iterations and would not benefit from additional rounds of training. 

\subsection{Inference}
After performing 10 EM iterations, we obtain reconstructions by taking the maximum likelihood word forms under the model. In the E-step, we sample $x^c \sim p(x^c \mid \{y_l^c: l\in L\})$, but now we want $x^c = \argmax p(x^c \mid \{y_l^c: l\in L\})$. We approximate this with an algorithm nearly identical to the E-step, except that we always select the highest probability candidate (instead of sampling) in Metropolis-Hastings iterations. 

\subsection{Underfitting the Model}
\label{sec:underfitting}
In prior work, models are trained to convergence in the M-step of EM. For example, the multinomial model of \citet{bouchard-etal-2007-probabilistic} has a closed-form MLE solution, and the log-linear model of \citet{bouchard-cote-etal-2009-improved} has a convex objective that is optimized with L-BFGS. In our experiments, we notice that training the neural model to convergence during M-steps will cause a degeneracy problem where reconstruction quality quickly plateaus and fails to improve over future EM iterations.

This degeneracy problem is crucially different from overfitting in the usual sense. In supervised learning, overfitting occurs when the model begins to fit spurious signals in the training data and deviates away from the true data distribution. On the other hand, precisely fitting the underlying distribution would cause our EM algorithm to get stuck -- if in a M-step the model fully learns the distribution from which samples were drawn, then the next E-step will draw samples from the same distribution, and the learning process stagnates. 

Our solution is to deliberately \textit{underfit} in the M-step. Intuitively, this gives more time for information to mix between the branches before the edit models converge to a common posterior distribution. We do this by training the model for only a small number of epochs in every M-step. We find that a fixed 5 epochs per step works well, which is far from the number of epochs needed for convergence. Our experiments in Section~\ref{sec:contextablation} show that this change significantly improves performance even when our model is restricted to the same local context as in \citet{bouchard-cote-etal-2009-improved}. 

\section{Experiments}
\label{sec:experiments}

\subsection{Comparison to Previous Models}
\label{sec:comparison}
We evaluate the performance of our model by computing the average edit distance between its outputs and gold Latin reconstructions. 

We experimented with several variations of the models used in prior work \citep{bouchard-etal-2007-probabilistic, NIPS2007_7ce3284b, bouchard-cote-etal-2009-improved} and chose the configuration which maximized performance on our dataset, referring to it as the \textit{classical} baseline. In particular, we found that extending the multinomial model in \citet{bouchard-etal-2007-probabilistic} to be conditioned on adjacent input characters and the previous output character as in \citet{bouchard-cote-etal-2009-improved} performed better than using the model from the latter directly, which used a log-linear parameterization. Given that we use an order of magnitude more data, we attribute this to the fact that the multinomial model is more flexible and does not suffer from a shortage of training data in our case. We confirm that this modified model outperforms \citet{bouchard-etal-2007-probabilistic, NIPS2007_7ce3284b} on the original dataset. For the learning algorithm, we keep the single sequence resampling algorithm from these papers. Although the more recent \citet{bouchard-cote-etal-2009-improved, doi:10.1073/pnas.1204678110} used ancestral resampling, the algorithm is focused on propagating information through large language trees, so it did not achieve a statistically significant improvement on the Romance languages, which only had a few nodes \citep{bouchard-cote-etal-2009-improved}. 

We also include an \textit{untrained} baseline to show how these methods compare to a model not trained with EM at all. The \textit{untrained} baseline evaluates the performance of a model initialized with fixed probabilities of self-substitutions, substitutions, insertions, and deletions, regardless of the context. We do not run any EM steps and take strings with the highest posterior probability under this model as reconstructions. We find that this baseline significantly outperforms the centroids baseline from previous work (4.88), so we use it as the new baseline in this work. 

\begin{figure*}[t!]
    \centering
   \includegraphics[width=\textwidth]{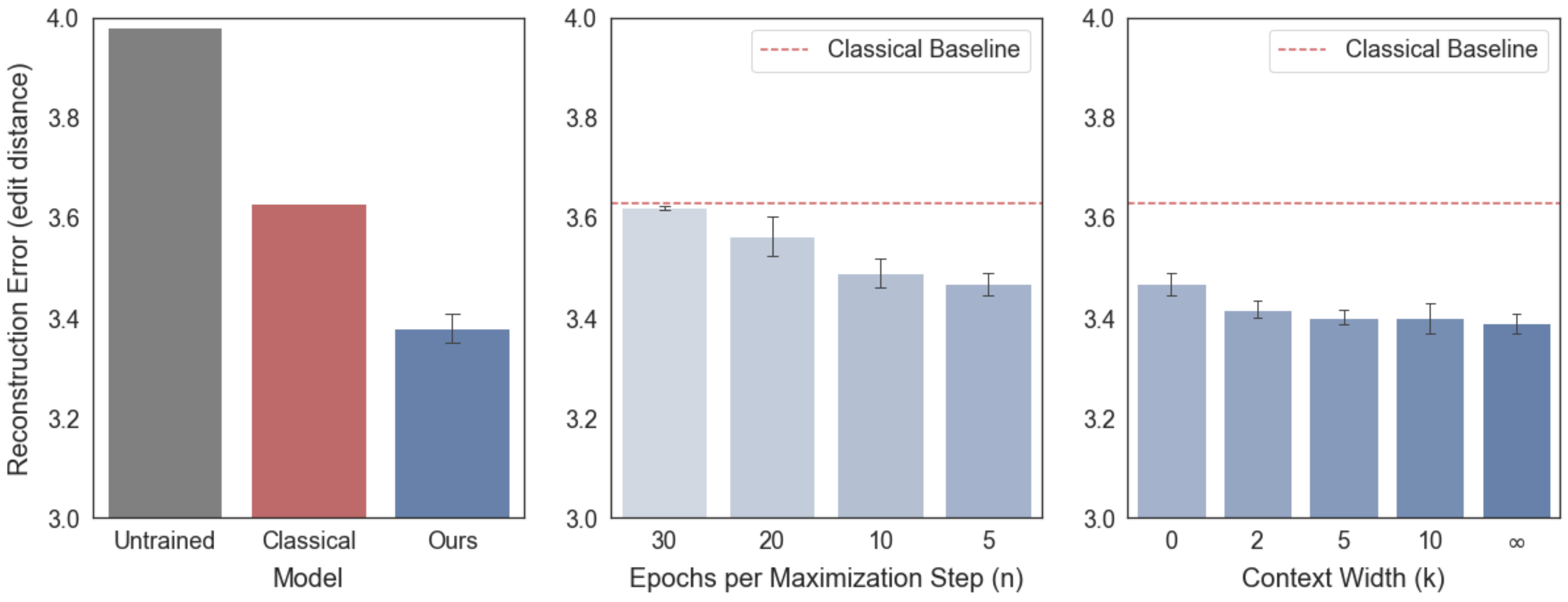}
    \caption{(Left) Our method significantly outperforms the classical baseline from  \citet{bouchard-cote-etal-2009-improved}. Although the improvement is only a 7\% reduction in terms of edit distance, we reduce the error rate by 70\% as much as the classical model did from an untrained baseline. (Middle) Reducing the number of epochs per maximization step underfits the model but results in better reconstructions in the long run. (Right) When the learning algorithm is well-regularized, conditioning edit probabilities on wider contexts results in more accurate reconstructions. 
    }
    \label{fig:results}
\end{figure*}

During training, we notice that different models take a different number of EM iterations to train, and some deteriorate in reconstruction quality if trained for too many iterations. Therefore, we trained all models for 10 EM iterations and report the quality of the best round of reconstructions in Figure~\ref{fig:results}. Since it may be impossible in practice to do early stopping without gold reconstructions, we also computed the final reconstruction quality for our models, but we observe only a minimal change in results ($\approx 0.02$ edit distance). Due to variance in the results, we report the mean and standard deviation across five runs of our method. 

\subsection{Ablation: Underfitting}
In this section, we describe an ablation experiment on the effect of under-training in the maximization step. Let $n$ represent the number of training epochs during each maximization step. Also, let $k$ represent the amount of context that our models have access to. When predicting an edit, the model can see $k$ characters to the left and right of the current input character (i.e., the window has length $2k+1$) and $k+1$ characters into the output history. Everything outside this range is masked out. Our standard model uses $n=5$ and $k=\infty$.

For this experiment, we set the context size to $k=0$ and run our method with $n \in \{5, 10, 20, 30\}$. The resulting reconstruction qualities are shown in Figure~\ref{fig:results}. Note that when $k=0$, our model is conditioned on the same information as that of \citet{bouchard-cote-etal-2009-improved}. When $n=30$, the model is effectively trained to convergence in every M-step. It completely fits the conditional distribution of edits in the samples, so it should learn the same probabilities as the multinomial model baseline. Indeed, the model with $n=30$ and $k=0$ achieves an edit distance of 3.61, which is very close to the 3.63 baseline. Given that this configuration is effectively equivalent to the classical method, we can incrementally observe the improvement from moving towards $n=5$ (our default). 

By reducing the number of epochs per maximization step ($n$), we observe a large improvement from 3.61 to 3.47. The general motivation for under-training the model was given in Section~\ref{sec:underfitting}. The remaining improvement comes from additional context, as we will demonstrate in the next subsection by moving towards $k=\infty$. 

\subsection{Ablation: Context Length}
\label{sec:contextablation}
In this section, we describe an ablation experiment on the effect of modeling longer contexts. Keeping $n=5$ fixed and using $k$ as defined in the previous subsection, we run our method three times for each of $k \in \{0, 2, 5, 10, \infty\}$ and report the average reconstruction quality in Figure~\ref{fig:results}.

Our results show that being able to model longer contexts does monotonically improve performance. The improvement is most drastic when expanding to a short context window ($k=2$). These findings are consistent with the knowledge that most (but not all) sound changes are either universal or conditioned only on nearby context \citep{campbell2013historical, Hock+2021}. With unlimited context length, our reconstruction quality reaches 3.38. Therefore, we attribute the overall improvement in our method to the changes of (1) modeling longer contexts and (2) underfitting edit models to learn more effectively with expectation-maximization.

\section{Discussion}

In this paper, we present a neural architecture and EM-based learning algorithm for the unsupervised reconstruction of protolanguage word forms. Given that previous work only modeled locally-conditioned sound changes, our approach is motivated by the fact that sound changes can be influenced by rich and sometimes non-local phonological contexts. Compared to modern sequence to sequence models, we also seek to regularize the hypothesis space and thus preserve the structure of character-level edits from classical models. 
On a dataset of Romance languages, our method achieves a significant improvement from previous methods, indicating that both richness and regularity are required in modeling phonological change.

We expect that more work will be required to scale our method to larger and qualitatively different language families. For example, the Austronesian language dataset of \citet{greenhill2008austronesian} contains order of magnitudes more modern languages (637 vs. 5) but significantly less words per language (224 vs. 3214) -- 
 efficiently propagating information across the large tree may be more important than training highly parameterized edit models in these settings. Indeed, \citet{bouchard-cote-etal-2009-improved, doi:10.1073/pnas.1204678110} produce high quality reconstructions on the Austronesian dataset by using ancestral resampling and sharing model parameters across branches. These improvements are not immediately compatible with our neural model; therefore, we leave it as future work to scale our method to settings like the Austronesian languages.

\section*{Limitations}
A major limitation of this work is that our method was designed for large cognate datasets with few languages. It may not be possible to train these highly parameterized edit models on datasets with more languages but fewer datapoints per language (e.g. the Austronesian dataset from \citet{greenhill2008austronesian}), and reconstruction in these datasets may benefit more from having efficient sampling algorithms and sharing parameters across branches \citep{bouchard-cote-etal-2009-improved}. Given the large amount of noise in the Romance language dataset, we also do not overcome the restriction in \citet{bouchard-etal-2007-probabilistic} of relying on a bigram language model of Latin. Moreover, inspecting learned sound changes is more difficult when using a neural model, so we leave a qualitative evaluation of unsupervised reconstructions from neural methods to future work. 

\bibliography{references}
\bibliographystyle{acl_natbib}

\appendix
\clearpage\newpage
\section{Appendix}
\label{sec:appendix}
\subsection{Dataset}
We describe the origin of our dataset and our preprocessing steps in Section~\ref{sec:dataset}. We show examples of some cognate sets in Table~\ref{ipa}, along with sample reconstructions from our best model.

\begin{table*}[h]
\centering
\begin{tabular}{llllll}
\hline \textbf{French} & \textbf{Italian} & \textbf{Spanish} & \textbf{Portuguese} & \textbf{Latin (Target)} & \textbf{Reconstruction} \\
\hline
 ablatif & ablativo & \textipa{aBlatiBo} & \textipa{5l5tivU} & \textipa{ablatIwUs} & \textipa{ablativU}\\
 \textipa{idKolik} & \textipa{draUliko} & \textipa{iDRauliko} & \textipa{id\textturnr aUlikU} & \textipa{hydraUlIkUs} & \textipa{idraUlikU} \\
 \textipa{inEfabl} & \textipa{ineffabile} & \textipa{inefaBle} & \textipa{in\textbari favEl} & \textipa{InEffabIlIs} & \textipa{inEfablE}\\
 \textipa{mAda} & mandato & mandato & \textipa{m5NdatUm} & \textipa{mandatUm} & \textipa{mandatU}\\
 \textipa{pKEsjO} & pessione & \textipa{pResjon} & \textipa{p\textturnr \textbari s5U} & \textipa{prEssIO} & \textipa{prEssO}\\ 
\textipa{pKOkKee} & prokreare & \textipa{pRokReaR} &  \textipa{p\textturnr uk\textturnr ia\textturnr} & \textipa{prOkrEarE} & \textipa{prOkrear} \\ 
\textipa{vokabylEK} & vokabolario & \textipa{bokaBulaRjo} &  \textipa{vuk5bulaRjU} & \textipa{wOkabUlarIUm} & \textipa{vokabylarEU} \\
 \textipa{ekonomi} & ekonomia & \textipa{ekonomia} &  \textipa{ekunumi5} & \textipa{OIkOnOmIa} & \textipa{ekunomia} \\
  \textipa{fekyl} & fekola & \textipa{fekula} &  \textipa{fEkul5} & \textipa{faIkUla} & \textipa{fEkyla} \\
    \textipa{lamine} & lamina & \textipa{lamina} &  \textipa{l5min5} & \textipa{lamIna} & \textipa{lamina} \\
  
\hline 
\end{tabular}
\caption{
IPA transcriptions for several cognate sets after our preprocessing steps, along with gold labels and example reconstructions from our best performing unsupervised reconstruction model}
\label{ipa}
\end{table*}

\subsection{Forward Dynamic Program}
\label{appendix:fdp}
The forward dynamic program computes the total probability of a output word form $p(y \mid x)$, marginalized over possible edit sequences $\Delta$. 
We first run inference with our neural models $q_\text{sub}$ and $q_\text{ins}$ to pre-compute the probabilities of all possible edits. For $i \in [\len(x)]$, $j \in [\len(y)]$, $op \in \{\text{sub}, \text{ins}, \text{del}, \text{end}\}$, let $C=(x, i, y[{:}j])$ be the context of the edit (and the input to the network). We compute:
\[
    \delta_{op}(i, j) \coloneqq
    \begin{cases}
        q_{op}(y[j] \mid C) & op \in \{\text{sub}, \text{ins}\}\\
        q_{\text{sub}}(\text{<del>} \mid C) & op=\text{del} \\
        q_{\text{ins}}(\text{<end>} \mid C) & op=\text{end} \\
    \end{cases}
\]
To compute the probability of editing $x$ into $y$, we define the subproblem $f_{op}(i, j)$ as the total probability of editing $x[{:}i]$ into $y[{:}j]$ such that the next operation is $op$. The recurrence can therefore be written as:
\[
\begin{split}
f_{\text{ins}}(i, j) = &\delta_{\text{ins}}(i, j-1) f_{\text{ins}}(i, j-1) \\& + \delta_{\text{sub}}(i, j-1) f_{\text{ins}}(i, j-1)
\end{split}\]
\[
\begin{split}
f_{\text{sub}}(i, j) = &\delta_{\text{end}}(i-1, j) f_{\text{ins}}(i-1, j) \\& + \delta_{\text{del}}(i-1, j) f_{\text{sub}}(i-1, j)
\end{split}\]
Which is in accordance with the dynamics described in Section \ref{sec:editproc}. The desired result is $p(y \mid x) =f_\text{sub}(\len(x), \len(y))$. We end on a substitution because it implies that the insertion for the final character has properly terminated. 

\subsection{Backward Dynamic Program}
\label{appendix:bdp}
The backward dynamic program computes the probability that an edit $(op, \omega, x, i, y')$ has occured, given the input string $x$ and output string $y$. We run the forward dynamic program first and use the notation $\delta$ and $f$ as defined in Appendix~\ref{appendix:fdp}. 

Define $g_{op}(i, j)$ as the posterior probability that the edit process has been in a state where the next operation is $op$ and it just edited $x[{:}i]$ into $y[{:}j]$. This is the same event as that of $f_{op}(i, j)$, but conditioned on the fact that the final output is $y$. The base case is therefore $g_{\text{sub}}(\len(x), \len(y))=1$. The dynamic program propagates probabilities backwards:
\[
\begin{split}
g_{\text{ins}}(i, j) = &\frac{\delta_{\text{ins}}(i, j)f_{\text{ins}}(i, j)}{f_{\text{ins}}(i, j+1)} g_{\text{ins}}(i, j+1)
\\& + \frac{\delta_{\text{end}}(i, j)f_{\text{ins}}(i, j)}{f_{\text{sub}}(i+1, j)}g_{\text{sub}}(i+1, j)
\end{split}\]
\[
\begin{split}
g_{\text{sub}}(i, j) = &\frac{\delta_{\text{sub}}(i, j)f_{\text{sub}}(i, j)}{f_{\text{ins}}(i, j+1)} g_{\text{ins}}(i, j+1)
\\& + \frac{\delta_{\text{del}}(i, j)f_{\text{sub}}(i, j)}{f_{\text{sub}}(i+1, j)}g_{\text{sub}}(i+1, j)
\end{split}\]
Essentially, each state receives probability mass from possible future states, weighed by its contribution in the forward probabilities. Finally, we recover the posterior probabilities of edits, denoted as $\delta'$:
\[\delta'_{\text{sub}}(i, j) = \frac{f_{\text{sub}}(i, j)g_{\text{ins}}(i, j+1)}{f_{\text{ins}}(i, j+1)} \delta_{\text{sub}}(i, j)\]
\[\delta'_{\text{ins}}(i, j) = \frac{f_{\text{ins}}(i, j)g_{\text{ins}}(i, j+1)}{f_{\text{ins}}(i, j+1)}\delta_{\text{ins}}(i, j) \]
\[\delta'_{\text{del}}(i, j) = \frac{f_{\text{sub}}(i, j)g_{\text{sub}}(i+1, j)}{f_{\text{sub}}(i+1, j)}\delta_{\text{del}}(i, j)\]
\[\delta'_{\text{end}}(i, j) = \frac{f_{\text{ins}}(i, j)g_{\text{sub}}(i+1, j)}{f_{\text{sub}}(i+1, j)}\delta_{\text{end}}(i, j)\]
Each $\delta'_{op}(i, j)$ corresponds to the same edit as $\delta_{op}(i, j)$, and so we obtain $p((op, \omega, x, i, y') \in \Delta \mid x, y)$ for all possible edits.

\subsection{Hyperparameters and Setup}
For our edit models, the input encoder is a bidirectional LSTM with $50$ input dimensions, $50$ hidden dimensions, and $1$ layer. The output encoder is a unidirectional LSTM with the same configuration. The dimension $50$ was found through a hyperparameter search over models of $d \in \{10, 25, 50, 100, 200\}$ dimensions. For training, we use the Adam optimizer with a fixed learning rate of $0.01$. All experiments were run on a single Quadro RTX 6000 GPU; however, GPU-based computations are not the bottleneck of our method. A single run of our standard method takes about $2$ hours. 


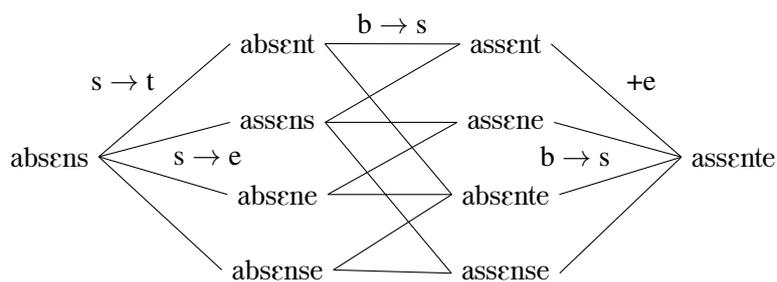
\begin{figure*}
\centering
\begin{tikzpicture}
\node [anchor=base](a) at (0, 0) {\textipa{absEns}}; 
\node [anchor=base](b1) at (3,1.5) {\textipa{absEnt}};
\node [anchor=base](b2) at (3,0.5) {\textipa{assEns}};
\node [anchor=base](b3) at (3,-0.5) {\textipa{absEne}};
\node [anchor=base](b4) at (3,-1.5) {\textipa{absEnse}};
\node [anchor=base](c1) at (6,1.5) {\textipa{assEnt}};
\node [anchor=base](c2) at (6,0.5) {\textipa{assEne}};
\node [anchor=base](c3) at (6,-0.5) {\textipa{absEnte}};
\node [anchor=base](c4) at (6,-1.5) {\textipa{assEnse}};
\node [anchor=base](d) at (9,0) {\textipa{assEnte}};
\draw (a.east) -- (b1.west) node[midway, above left]{s $\rightarrow$ t};
\draw (a.east) -- (b2.west);
\draw (a.east) -- (b3.west) node[midway, above right]{s $\rightarrow$ e};
\draw (a.east) -- (b4.west);
\draw (b1.east) -- (c1.west) node[midway, above]{b $\rightarrow$ s};
\draw (b2.east) -- (c2.west);
\draw (b3.east) -- (c3.west);
\draw (b4.east) -- (c4.west);
\draw (b1.east) -- (c3.west);
\draw (b2.east) -- (c4.west);
\draw (b2.east) -- (c1.west);
\draw (b3.east) -- (c2.west);
\draw (b4.east) -- (c3.west);
\draw (c1.east) -- (d.west) node[midway, above right]{+e};
\draw (c2.east) -- (d.west);
\draw (c3.east) -- (d.west) node[midway, above left]{b $\rightarrow $ s};
\draw (c4.east) -- (d.west);
\end{tikzpicture}
\caption{Example of possible proposals when the current reconstruction is \textit{\textipa{absEns}}, and it has a modern cognate \textit{\textipa{assEnte}}. We only show edit paths between the current sample and its Italian cognate here, but candidates can also be on paths between the current sample and its other modern cognates.}
\label{fig:mineditpath}
\end{figure*}

\end{document}